%% file: main.tex
\documentclass{article}

\usepackage{times}
\usepackage{ijcai18}

\usepackage{siddmath}
\usepackage[ruled,vlined]{algorithm2e}
\usepackage{multirow}
\usepackage{enumitem}
\usepackage{hhline}
\usepackage{paralist}
\usepackage{fancyhdr}
\usepackage{natbib}
\usepackage{xcolor}
\usepackage[small]{caption}

\newcommand{\?}{\raisebox{.5pt}{\textcircled{\raisebox{-.9pt} {\small ?}}}}

\title{An Anytime Algorithm for Task and Motion MDPs} 
\author{Siddharth Srivastava\thanks{Some of the work was done while this author as at United Technologies Research Center}$^\ddagger$, Nishant Desai$^\dagger$, Richard Freedman$^\mathsection$, Shlomo Zilberstein$^\mathsection$  \\
  $^\ddagger$Arizona State University, $^\dagger$United Technologies
  Research Center, $^\mathsection$University of Massachusetts}

\begin{document}
\maketitle

\begin{abstract}
  Integrated task and motion planning has emerged as a challenging
  problem in sequential decision making, where a robot needs to
  compute high-level strategy and low-level motion plans for solving
  complex tasks. While high-level strategies require decision making
  over longer time-horizons and scales, their feasibility depends on
  low-level constraints based upon the geometries and continuous
  dynamics of the environment. The hybrid nature of this problem makes
  it difficult to scale; most existing approaches focus on
  deterministic, fully observable scenarios.  We present a new
  approach where the high-level decision problem occurs in a
  stochastic setting and can be modeled as a Markov decision
  process. In contrast to prior efforts, we show that complete MDP
  policies, or contingent behaviors, can be computed effectively in an
  anytime fashion. Our algorithm continuously improves the quality of
  the solution and is guaranteed to be probabilistically complete. We
  evaluate the performance of our approach on a challenging, realistic
  test problem: autonomous aircraft inspection. Our results show that
  we can effectively compute consistent task and motion policies for
  the most likely execution-time outcomes using only a fraction of the
  computation required to develop the complete task and motion
  policy.
\end{abstract}

\input{intro}

\input{background}

\input{formal}

\input{alg}

\input{empirical}

\input{related}

\input{conclusions}

\section*{Acknowledgements}
This  material is based upon work supported by the Defense Advanced
  Research Projects Agency (DARPA) and Space and Naval Warfare Systems
  Center Pacific (SSC Pacific) under Contract
  No. N66001-16-C-4050. Any opinions, findings and conclusions or
  recommendations expressed in this material are those of the authors
  and do not necessarily reflect the views of the DARPA or SSC
  Pacific.
\bibliographystyle{named}
\bibliography{planning}

\end{document}

%% file: intro.tex
\section{Introduction}

\begin{figure}
  \includegraphics[height=1in]{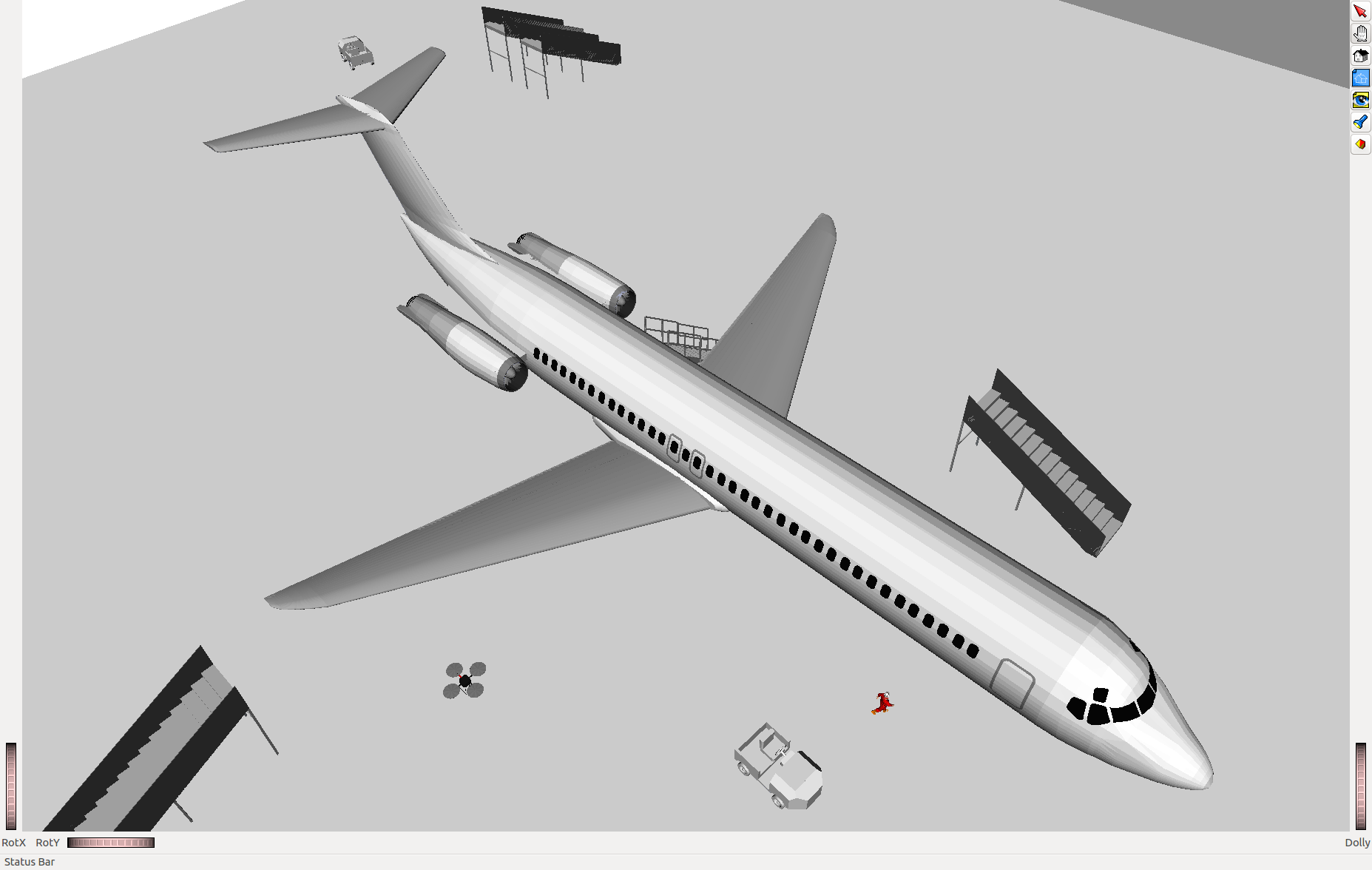}
  \includegraphics[height=1in]{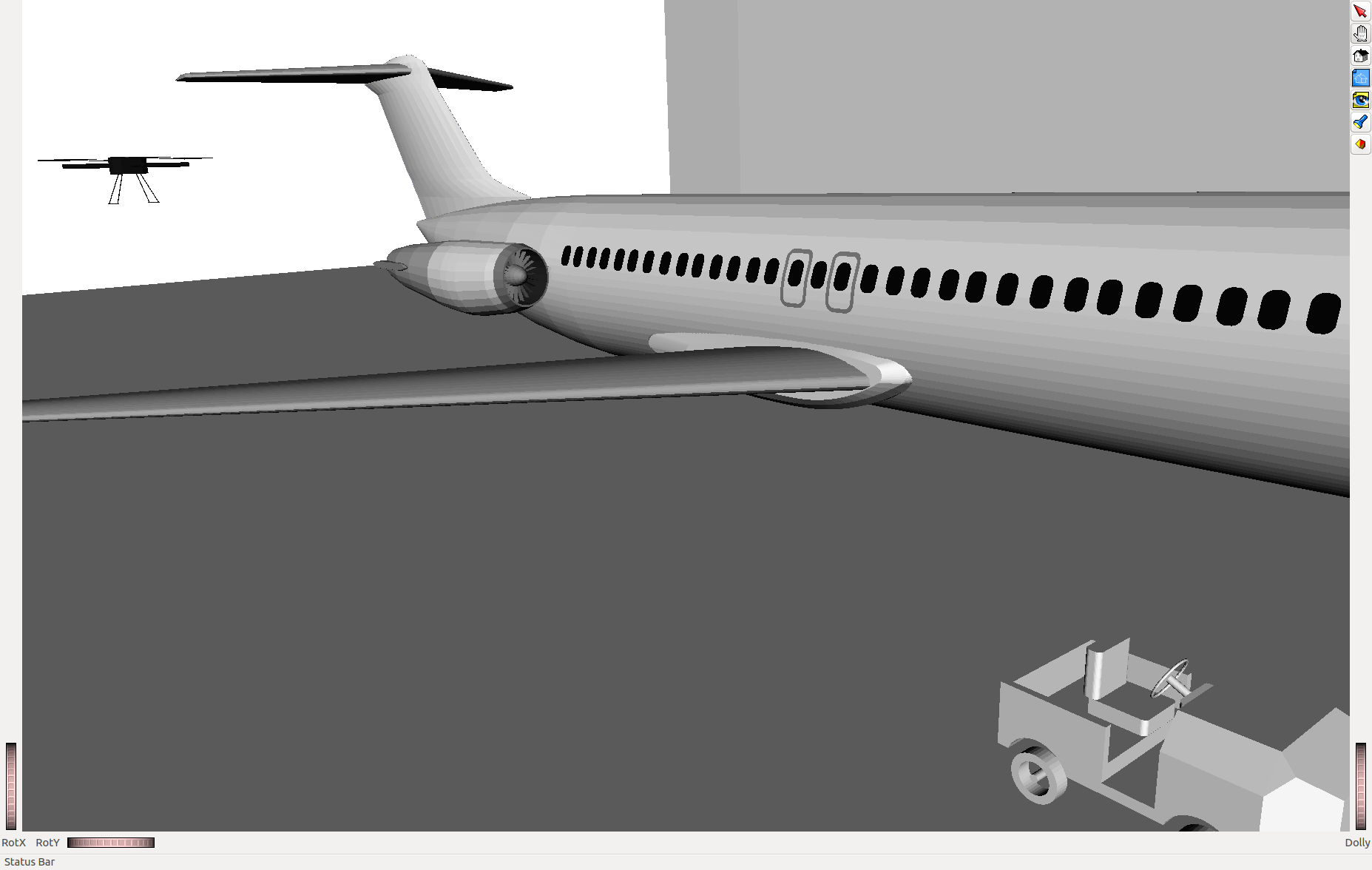}
\caption{The aircraft inspection scenario}\label{fig:scenario}
\end{figure}

In order to be truly helpful, robots will need to be able to accept commands from humans at high-levels of abstraction, and autonomously execute them.
Consider the problem of inspecting an aircraft (Fig.\,\ref{fig:scenario}). 
In order to autonomously plan and execute such a task, the robots (UAVs in this case) will need to be able to make  high-level inspection decisions on their own, while satisfying low-level constraints that arise from environment geometries and the limited capabilities of the UAVs. High-level decisions can include selecting where to go next, with whom to communicate, and what to inspect.  These decisions need to take into account the uncertainty in the UAV's actions.

For instance, at the start of an aircraft's inspection, one may know that the left wing has a structural problem, but the location of the fault may not be known precisely. When a UAV inspects the left wing, its sensors may succeed with probability 0.9, and so on. In order to solve this task autonomously, the UAV needs to select  which pose to fly to next, which trajectory to use in order to do so, and  the order in which to carry out inspections while making sure that it always has sufficient battery to return to the docking station and that it does not collide with any object in the environment. The feasibility of a high-level strategy for inspection therefore depends on the battery power required for each high-level operation such as ``move to left wing"; ``inspect left wing", etc., which in turn depends on the low-level motion plan selected, which in turn depends on the hangar's geometric layout and the physical geometry of the UAV. Throughout this paper, we will use the term ``high-level'' to represent a discrete MDP and ``low-level'' to refer to a motion planning problem.

The framework of Markov decision processes (MDPs) can express discrete sequential decision making (SDM) problems. Numerous advances have been made in solving MDPs~\citep{russell15_robust}. However, the scalability of these approaches relies upon a few key properties, including a bounded branching factor (or the set of possible actions) and the ability to express a problem accurately using discrete state variables. Both of these properties fail to hold in problems such as those described above. Recent work on deterministic, integrated task and motion planning~\citep{kaelbling11_hierarchical,erdem11_tmp,srivastava14_tmp,dantam16_incremental} shows that hierarchical approaches are useful for such problems. 

Computing task and motion policies for MDPs presents a new set of challenges not encountered in computing  task and motion  plans for deterministic scenarios. In particular, selecting an action for a state while ensuring a feasible refinement requires knowing the history of actions used to reach that state, since effects on properties that were abstracted away (such as battery usage) cannot be modeled accurately  at the high level. A direct application of classical task and motion planning techniques  is further limited by the number of possible high-level action paths that can be taken during an execution. Indeed, the task and motion planning literature makes it clear that computing a single high-level sequence of actions that is feasible with low-level constraints is a challenge; the extension to MDPs expands the problem to computing a feasible high-level sequence of actions for every possible stochastic outcome of a high-level action. 

In this paper, we investigate the problem of computing task and motion policies and show that principles of abstraction can be used to effectively model the problem, as well as to solve it by dynamically refining the abstraction used. We address the problem of computational complexity by developing an anytime algorithm that rapidly produces feasible policies for a high likelihood of scenarios that may be encountered during execution. Our methods can therefore be used to start the execution before the complete problem is solved; computation could continue during execution. The continual policy computation reports the probability of encountering situations which have not been resolved yet. This can be used to select the point at which execution is started in a manner appropriate to the application. In the worst case, if an unlikely event is encountered before the ongoing policy computation resolves it, execution could be brought to a safe state; in situations where this is not possible, one could wait for the entire policy to be computed with motion plans. In this way our approach offers a trade-off between pre-execution guarantees and pre-computation time requirements.

The rest of this paper is organized as follows. Sec.\,\ref{sec:background} introduces the main concepts that we draw upon from prior work. Sec.\,\ref{sec:formal} presents our formalization of abstractions and representations. This is followed by a description of our algorithms (Sec.\,\ref{sec:alg}). Sec.\,\ref{sec:empirical} presents an empirical evaluation of our approach in a test scenario that we created using open-source 3D models of aircraft and various hangar components. Sec.\,\ref{sec:related} discusses the relationship of the presented work and contributions with prior work.

%% file: background.tex
\section{Background}
\label{sec:background}
A Markov decision process (MDP) $\langle S, A, T, R, \gamma \rangle$
is defined by a set of states $S$, a set of actions $A$, a transition
function $T:S\times A \rightarrow \mu S$ that gives the probability
distribution over result states upon the application of an action on a
state; a reward function $R: S \rightarrow \mathbb{R}$; and a
discounting factor $\gamma \le 1$. We will use $T(s,a)$ as a function that maps a state to its probability. 
We will be particularly interested in
MDPs with absorbing states and $\gamma=1$, or, stochastic shortest path
problems~\citep{bertsekas91_ssp}. In this class of MDPs, the reward
function yields negative values (action costs) for states except the
absorbing states $G$. Absorbing states give zero rewards; once the
agent reaches an absorbing state, it stays in it: $\forall a \in A,
g\in G, R(g)=0; T(g,a)(g)=1$. We will consider SSPs that have a
known initial state $s_0$ and a finite time horizon, $h$, which
represents an uppper bound on the number of discrete decision making
steps available to the agent.

Solutions to MDPs are represented as policies. A policy $\pi: S \times
\set{1,\ldots, h} \rightarrow A$ maps a state and the timestep at
which it is encountered, to the action that the agent should execute
while following $\pi$. Given an MDP, the optimal ``policy" of the agent is defined as one that maximizes the expected long-term reward $\sum_{i=1}^h r_i$, where $r_i$ is the reward obtained at timestep $i$ following the function $R$. Our notion of policies includes non-stationary policies since the optimal policy in a finite horizon MDP need not be stationary. In principle, dynamic programming can be used to compute the optimal policy in this setting just as in the infinite horizon setting:

\begin{eqnarray}
V^0(s) &=& R(s)\\
V^i(s) &=& R(s) + max_a \sum_{s'} T(s,a)(s')V^{i-1}(s')
\end{eqnarray}

Here $V^i$ is the $i$-step-to-go value function. Since we are given the initial state $s_0$,
non-stationary policies can be expressed as finite state machines (FSMs). We will
consider policies that are represented as tree-structured FSMs, also
known as contingent plans. Several algorithms have been developed to
solve SSPs. The LAO* algorithm~\citep{hansen01_lao} was developed to
incorporate heuristics while computing solution policies for SSPs. 
\cite{kolobov11_ssp} developed general methods for solving SSPs in
the presence of dead-ends.

Specifying real-world sequential decision making problems as MDPs
using explicitly enumerated state lists usually results in large,
unweildy formulations that are difficult to modify, maintain, or
understand. Lifted, or parameterized representations for MDPs such as
FOMDPs~\citep{sanner09_fomdp}, RDDL ~\citep{sanner10_rddl} and
PPDDL~\citep{younes04_ppddl} have been developed for overcoming these
limitations. Such languages separate an MDP \emph{domain},
constituting parameterized actions, functions and predicate
vocabularies, from an MDP \emph{problem}, which expresses the
specific objects of each type and a reward function. We refer to
\cite{helmert09_pddl_grounding} for a general introduction to these
concepts. W.l.o.g, we consider the vocabulary to consist of predicates
alone, since functions can be represented as special predicates.  A
grounded predicate is a predicate whose parameters have been
substituted by the objects in an MDP problem. For instance, Boolean
valuations of the grounded predicate \emph{faultLocated(LeftWing)}
express whether 
the LeftWing's fault's precise location was 
identified.
In our framework, states are defined as valuations of grounded
predicates in a given problem.  Although this framework usually
expresses \emph{discrete} properties, it 
can be extended
naturally to model actions that have continuous action arguments and
depend on and affect geometric properties of the environment.

\begin{figure}[t]
        \begin{small}
                \begin{tabular}{r|p{4in}}\multicolumn{2}{p{4in}}
                  {Action:
                    \emph{inspect(Structure $s$, Trajectory $tr$)}}\\
                  precond & $\emph{batterySufficient}(tr) \land $  \emph{inspects}($tr, s$)$\land$ collisionFree($tr$)\\
                  effect & \emph{faultLocated}$(s) \qquad 0.8$ \\
                  & $\lnot$\emph{faultLocated}$(s) \qquad 0.2$ \\
                  & \emph{decrease(batteryLevel($c(tr))$)}\\
                \end{tabular}
        \end{small}
        \caption{Specification of a  stochastic action model}\label{fig:inspect_concrete}
\end{figure}

\begin{example}
  \label{eg:inspect_concrete}
  Fig.\,\ref{fig:inspect_concrete} shows the specification for an
  \emph{inspect} action in the aircraft inspection domain in a
  language similar to PPDDL (some syntactic elements have been
  simplified for readability).  This action models the act of
  inspecting a structure $s$ while following the path $tr$. We use \emph{batterySufficient}$(tr)$ as an
  abbreviation for \emph{batteryRemaining}$ -
  $\emph{batteryRequired($tr$)}. Intuitively, the specification states that if this
  action is executed in a state where the  battery is sufficient and the selected trajectory satisfies constraints for being an inspection trajectory (the
  \emph{precondition} is satisfied),  it will result in locating
  the fault with the probability 0.8. In any case, the battery's
  charge will be depleted by an amount depending on the trajectory
  used for inspection $c(tr)$.  The \emph{inspects(c, tr)} predicate is true if the trajectory $tr$ ``covers" the given structure. Different interpretations for such predicates would result in different classes of coverage patterns.
\end{example}

%% file: formal.tex
\section{Formal Framework}
\label{sec:formal}
Let $X$ be a set of states and $S$ a set of abstract states. We define
a \textbf{state abstraction} as a surjective function $\alpha:
X\rightarrow S$. We focus on \emph{predicate
  abstractions}, where the abstraction function effectively projects
the state space into a space without a specified set of
predicates. Given a set of predicates $\mathcal{P}$ that are retained
by a predicate abstraction, the states of the abstract state space
are equivalence classes defined by the equivalence relation $s_1 \sim
s_2$ iff $s_1$ and $s_2$ agree on the valuations of every 
predicate in $\mathcal{P}$, grounded using the objects in the problem. 

For any $s\in S$, the \emph{concretization
  function} $\gamma_\alpha(s) = \set{x\in X: \alpha(x)=s}$ denotes the
set of concrete states \emph{represented by the abstract state}
$s$. For a set $C\subseteq X$, $[C]_\alpha$ denotes the smallest set
of abstract states representing $C$. Generating the complete
concretization of an abstract state can be computationally
intractable, especially in cases where the concrete state space is
continuous and the abstract state space is discrete. In such
situations, the concretization operation can be implemented as a
\emph{generator} that incrementally computes or samples elements
from an abstract state's concretization.

	
\textbf{Action abstraction functions} can be defined similarly. The
main form of an action abstraction function is to drop action
arguments, which leads to predicate abstractions to eliminate all
predicates that used the dropped arguments in the action's
description. This process can also model non-recursive temporal
abstractions since a macro or a high-level action with multiple
implementations~\citep{marthi07_angelic} can be modeled as an action whose arguments include
the arguments of its possible implementations as well as an auxiliary
argument for selecting the implementation. The concretization
  of an action abstraction function is the set of actions
corresponding to different instantiations of the dropped action
arguments. Concretization functions for action abstraction functions
can also be implemented as generators.

Formally, the concretization of each high-level action corresponds to a set of motion planning problems. We will use the notation $a(x_1\mapsto o_1)$ to denote a grounded action, whose $x_1$ argument has been instantiated with the element $o_1$ defined by the underlying MDP problem (Sec.\,\ref{sec:background}). Let $a(\bar{x}, \bar{y})$ be a concrete action where $\bar{x}$ ($\bar{y}$) are ordered, typed discrete (continuous) arguments. The \emph{concretization of the instantiated abstract action} $\gamma([a](\bar{x}\mapsto \bar{o}))$ is the set of actions $\set{a(\bar{x}\mapsto \bar{o}, \bar{y}\mapsto \bar{o'}): \bar{o'} \emph{ is a tuple of elements with types and arity specified by  } y}$. Predicates in action preconditions specify the constraints that these arguments need to satisfy. Common examples for continuous arguments include robot poses and motion plans;  predicates about them may include \emph{collisionFree($tr$)}, which is true exactly when the trajectory $tr$ has no collisions as well as \emph{inspects} (Eg.~\ref{eg:inspect_concrete}).

Both state and action abstractions affect the transition function of
the MDP. The actual transition probabilities of an abstract MDP depend
on the policy being used and are therefore difficult to estimate
accurately~\citep{bai16_markovian,li06_abstractMDP,singh95_abstractRL}. In
this paper, we will use an optimistic estimate of the true transition
probabilities when expressing the abstract MDP. Such estimates are
related to upper bounds for reachability used in prior
approaches for
reasoning in the presence of hierarchical abstractions~(e.g., \citep{marthi07_angelic,haddawy96_drips}).

\begin{example}

Consider the action presented in Eg.\,\ref{eg:inspect_concrete}
Such actions are difficult to plan with however, since the $tr$
argument is a high-dimensional real-valued vector. We can abstract
away this argument to construct the following abstraction:

        \begin{small}
                \begin{tabular}{r|p{4in}}\multicolumn{2}{p{4in}}
                  {Action:
                    \emph{[inspect](Structure $s$)}}\\
                  precond & $\emph{batterySufficient}$\\
                  effect & \emph{faultLocated}$(s) \qquad 0.8$ \\
                  & $\lnot$\emph{faultLocated}$(s) \qquad 0.2$ \\
                  & $\?$\{\emph{batteryLevel, batterySufficient}\}
                \end{tabular}
        \end{small}

Dropping the $tr$ argument from each predicate that results in
abstract predicates of lower arities. The zero-arity
\emph{batterySufficient} becomes a Boolean state variable and
\emph{batteryLevel} becomes a numeric variable. The symbol $\?$
indicates that this action affects the predicates \emph{batteryLevel}
and \emph{batterySufficient}, but its effects on these predicates
cannot be determined due to abstraction.

An optimistic representation of this abstract action would state that
it does not reduce \emph{batteryLevel} and consequently, does
not make \emph{batterySufficient} false.
\end{example}

This approach for abstraction is computationally better than a high-level representation that discretizes the continuous variables, as it does not require the addition of constants representing discrete pose or trajectory names to the vocabulary. This is desirable because the size of the state space would be exponential in the number of such discretized values that are included.

%% file: alg.tex
\section{Overall Algorithmic Framework}
\label{sec:alg}
\begin{algorithm}[t]
\begin{small}
\SetKwFunction{estimatePathCosts}{estimatePathCosts} \SetKwFunction{ancestors}{ancestors} \SetKwFunction{refinePath}{refinePath}
\SetKwFunction{randomUniform}{randomUniform} \SetKwFunction{computeProportionRefined}{computeProportionRefined}
\SetKwData{None}{None}
\KwData{domain $\mathcal{D}$, problem $\mathcal{P}$, motionPlanner $MP$,  SSP Solver $SSP$}
\KwIn{threshold $t$}
\KwResult{Task and motion policy for $\langle \mathcal{D}, \mathcal{P}\rangle$}
\nl policyTree $\leftarrow$ $SSP$.getContingentPlan($\mathcal{P}$.$\vec{f_{0}}$, $\mathcal{D}$, $\mathcal{P}$)\; 
\nl currentState $\leftarrow$ $\mathcal{P}$.$\vec{f_{0}}$; proportionRefined $\leftarrow$ 0.0; replanBias $\leftarrow$ 0.5\; 
\nl partialTraj $\leftarrow$ \None\;
\nl leafQueue $\leftarrow$ \estimatePathCosts{policyTree, partialTraj}\;
\nl \While{resource limit not reached {\bf and} leafQueue.size() $\neq$ 0 {\bf and} proportionRefined $<$ $t$}
{
   \nl  pathToRefine $\leftarrow$ \ancestors{leafQueue.pop()}\;
   \nl  \While{resource limit not reached {\bf and} pathToRefine.length() $\neq$ 0}
    {
       \nl  (success, partialPathTraj, failureNode, failureReason) $\leftarrow$ \refinePath{pathToRefine, partialTraj, policyTree, $MP$}\;
       \nl  \If{{\bf not} success {\bf and} failureReason $\ne$ \None} {
           \nl  policyTree $\leftarrow$ $SSP$.replan(failureNode, failureReason)\;
           \nl  {\bf break}\;
        } \Else {\nl \If{ not success}{
            \nl \For{node $\in$ partialPathTraj} {
                \nl partialTraj[node] $\leftarrow$ partialPathTraj[node]
            	}
            }
        }
    }
    \nl leafQueue $\leftarrow$ \estimatePathCosts{policyTree, partialTraj}\;
    \nl proportionRefined $\leftarrow$ \computeProportionRefined{policyTree, partialTraj}
}   

\caption{\small Anytime Task and Motion MDP (ATM-MDP)}\label{alg:overall}
\end{small}
\end{algorithm}

The ATM-MDP algorithm (Alg.\,\ref{alg:overall}) presents the main outer loop of our approach for computing a task and motion policy. It assumes the availability of an SSP solver that can generate tree-structured policies (starting at a given initial state) for solving an SSP, a motion planner for refinement of actions within the policy, and a module that determines the reason for infeasibility of a given  motion planning problem. The overall algorithm operates on  root-to-leaf paths in the SSP solution.

The main computational problem is that the number of possible paths to refine grows exponentially with the time horizon. Waiting for a complete refinement would result in a lot of wasted time as most paths may correspond to outcomes that are unlikely to be encountered. Every path is associated with the probability $p$ that an execution would follow that path; and a cost $c$ of refining that pat. Ideally, we would like to compute an ordering of these paths so that at every time instant, we compute as many of the most likely paths as can be computed up to that time instant. Unfortunately, achieving this would be infeasible as it would require solving multiple \emph{knapsack} problems. Instead, we order the paths by the ratio $p/c$ for refinement (lines 4-15). 

\begin{theorem} Let $t$ be the time since the start of the algorithm at which the refinement of any root-to-leaf path is completed. If path costs are accurate and constant  then the total probability of unrefined paths at time $t$ is at most $1 - opt(t)/2$, where $opt(t)$ is the best possible refinement (in terms of the probability of outcomes covered) that could have been achieved in time $t$.
\end{theorem}
The proof follows from the fact that the greedy algorithm achieves a 2-approximation for the knapsack problem. In practice, the true cost of refining a path cannot be determined prior to refinement. We therefore estimate the cost as the product of the parameter ranges covered by the generator of each action in the path. This results in lower bounds on the ratios $p/c$ modulo constant factors, since a path could be refined before all the generator ranges are exhausted. In this way it doesn't over-estimate the relative value of refining a path. As we show in the empirical section, the resulting algorithm yields the concave performance profiles desired of anytime algorithms.


The \emph{while} loop iterates over these paths while recomputing the priority queue keys after each iteration. Within each iteration, the algorithm tries to compute a full motion planning refinement of the path. First, the entire path (\emph{pathToRefine}) is extracted from the leaf (line 6). 
The \emph{refinePath} subroutine attempts to find a motion planning refinement (concretization) for \emph{pathToRefine}. If it is unable to find a complete refinement for this path, it either (a) returns with a reason for failure along with a partial trajectory going up to the deepest node in the path for which it was able to compute a feasible motion plan, or (b) backtracks to return a partial trajectory that will result in a future refinePath call for a parent node of a node for which a motion planning refinement couldn't be found. 

\begin{algorithm}[t]
\begin{small}
\SetKwFunction{head}{head} \SetKwFunction{extractPose}{extractPose} \SetKwFunction{targetPoseGen}{targetPoseGen} 
\SetKwFunction{getNext}{getNext} \SetKwFunction{GetMotionPlan}{GetMotionPlan}
\SetKwData{None}{None} \SetKwData{False}{False} \SetKwData{True}{True}
\KwIn{pathToRefine, partialTraj, policyTree, motionPlanner}
\KwOut{success: indicator of successful refinement; partialPathTraj: refined path up to the first failure; failureNode, failureReason: failure information}
\nl node $\leftarrow$ \head{pathToRefine}; partialPathTraj $\leftarrow$ \None\;
\nl \For{node $\in$ pathToRefine} {
    \nl a $\leftarrow$ policyTree[node]\;
    \nl \eIf{partialTraj = \None} {
        \nl $pose_1$ $\leftarrow$ InitialPose\;  
    } {\nl $pose_1$ $\leftarrow$ \extractPose{partialTraj[parent(node)]}\;
	\nl \While{resource limit not reached {\bf and} partialPathTraj[node] = \None}{    
    \nl $pose_2$ = \targetPoseGen{a}\;
    \nl \If{\GetMotionPlan{$pose_1$, $pose_2$} succeeds} {
        \nl partialPathTraj[node] $\leftarrow$ ComputePath\;
       \nl {\bf break}\;
    } }
    \nl \If{partialPathTraj[node] = \None}{
    	\nl \eIf{Bernoulli(replanBias).sample()}
    		{
        		\nl return (\False, partialPathTraj, node, FailureReason)\;}
        	{
        		\nl partialPathTraj.remove(node.parent())\;
        		\nl return (\False, partialPathTraj, node.parent(), None )
        	}
    }
}}
return (\True, partialPathTraj, \None, \None)\;
\caption{Subroutine refinePath}\label{alg:refine}
\end{small}
\end{algorithm}

For partial trajectories under (a) (line 9), Alg.\,\ref{alg:overall} calls an SSP solver after adjusting its initial state and domain definitions to include the \emph{FailureReason}. The policy computed by the SSP solver is then merged with the existing policy and the while loop continues. For partial trajectories  along case (b) (line 12), the path is added back to the queue with a partial, successful trajectory that results in backtracking. 

If \emph{refinePath} is successful in computing a full refinement, the while loop continues with an updated priority queue. 
In each iteration of the while loop, we compute the total probability of refined paths -- this probability gives us the likelihood of being able to successfully execute the policy in its current state of refinement.

The \emph{refinePath} subroutine (Alg.\,\ref{alg:refine}) attempts to compute a motion plan for each action in a given path. 
More precisely,  it uses a generator to sample the possible concretizations for each action and test their feasibility. A feasible solution to any one of these motion planning problems is considered a feasible refinement of that abstract action. \emph{refinePath} starts by selecting the first node in the path that needs to be refined in line 1 (Alg.\,\ref{alg:overall} may result in situations where a prefix of a path has already been refined by a prior call to \emph{refinePath}, due to line 14 in that algorithm).

It then iterates over possible target poses for the selected action (lines 8 through 11). If a feasible motion plan is found, then the algorithm refines the next action in the path. If not, it stochastically chooses to either re-invoke the SSP by returning a \emph{FailureReason}, or to backtrack by invalidating the current node's path (line 15) by removing it from \emph{partialPathTraj} and returning to follow lines 12-13 in Alg.\,\ref{alg:overall}.

\begin{figure}[t]
\centering
\includegraphics[height=1in]{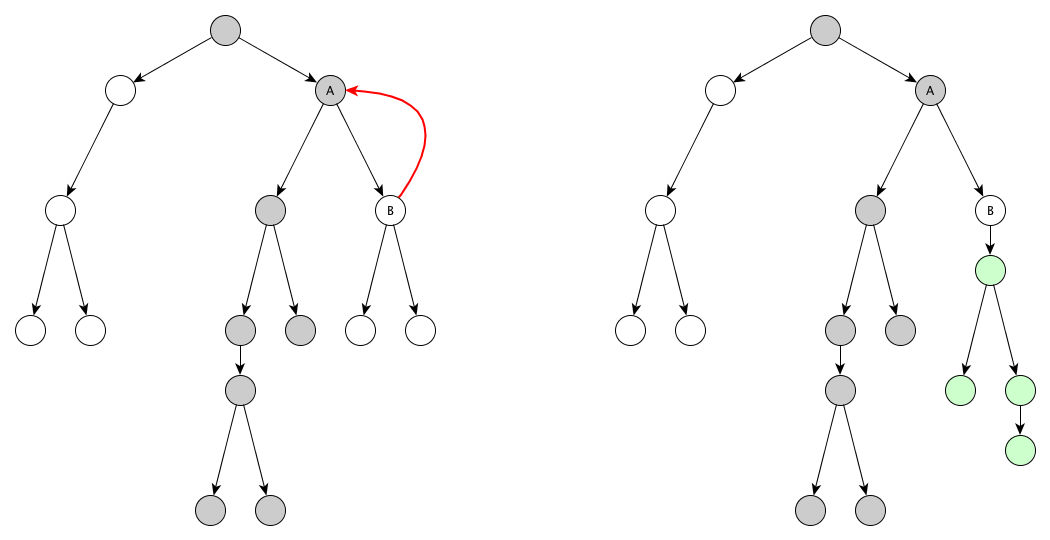}
\caption{Left: Backtracking from node B invalidates the subtree rooted at A. In doing so, the work done in refining the node A's left child, in gray, is lost. Right: In some cases, replanning from node B requires less work than re-refining the invalidated subtree.}
\label{fig:backtracking}
\end{figure}

Though a backtracking search through all possible motion plans is required to guarantee the completeness of the algorithm, we find in practice that replanning with a new initial state and replacing the subtree rooted at a failed node with a new SSP solution is often more time efficient. This is because backtracking to an ancestor of the failed node invalidates the motion plans associated with all paths passing through that ancestor, often causing a large amount of previously completed work to be thrown out. This situation is illustrated in Figure \ref{fig:backtracking}. For this reason, we stochastically choose between backtracking and replanning and settle for \emph{probabilistic} completeness of the search algorithm.

\paragraph{Properties of the Algorithm}
Our algorithm solves the dual problems of synthesizing a strategy as well as computing motion plans while ensuring that the computed strategy has a feasible motion plan.  It factors a hybrid planning problem into a succession of discrete SSPs and motion planning problems. The algorithm can compute solutions even when most discrete strategies have no feasible refinements. A few additional salient features of the algorithm are:
\items{
\item The representational mechanisms for encoding SSPs do not require discretization, thus providing scalability.
\item The SSP model dynamically improves as the motion planning problems reveal errors in the high-level model in terms of \emph{FailureReason}s.
\item Prioritizing paths of relative value gives the algorithm a desirable anytime performance profile. This is further evaluated in the empirical section.
}




%% file: empirical.tex
\section{Empirical Evaluation}
\label{sec:empirical}
\begin{figure*}[t]
  \centering
  \includegraphics[width=2.2in]{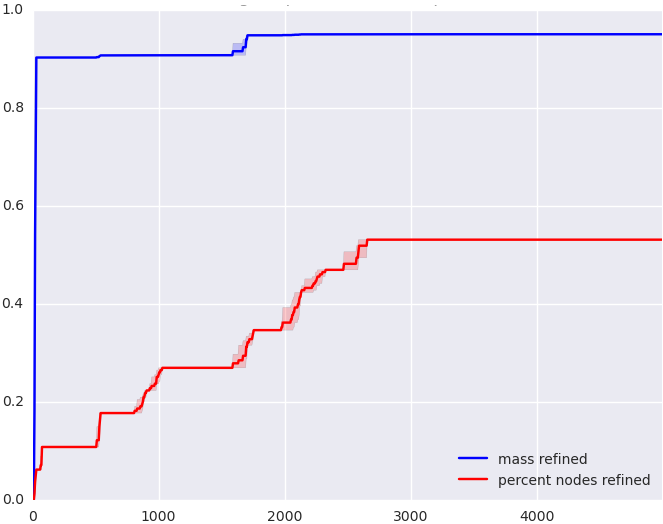}
  \includegraphics[width=2.2in]{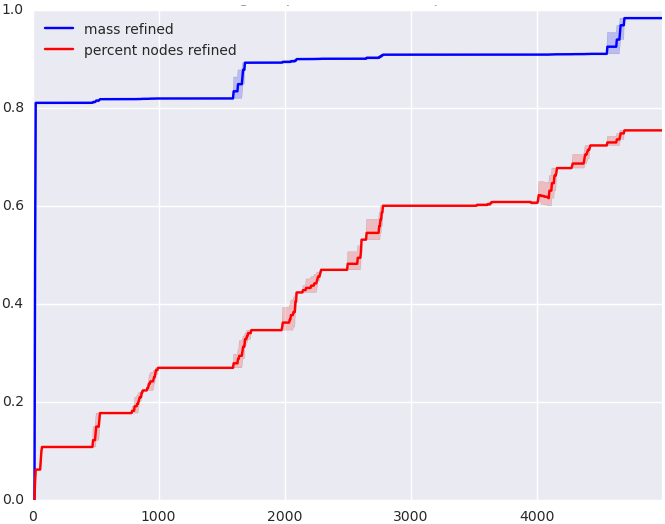}
  \includegraphics[width=2.2in]{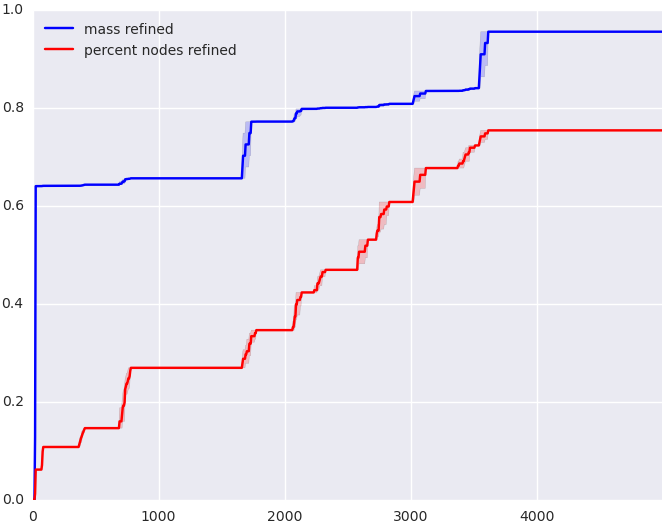}
\caption{\small Performance of our anytime algorithm for solving MDPs using
  dynamic abstractions. The plots from left to right corresponds to
  formulation of the problem with 5\%, 10\%, and 20\% rates of failure
  of the abstract actions described in the text. The blue lines (red
  lines) plot the probability mass of possible outcomes (proportion of
  nodes in the policy graph) that is covered by the partially computed
  policy as computation time (x axis, in seconds) evolves. }\label{fig:results_mdp}
\end{figure*}

We implemented the algorithms presented in Sec.\,\ref{sec:alg} using an implementation of LAO*~\citep{hansen01_lao} as the SSP solver. We used the OpenRAVE~\citep{diankov10_openrave} system for modeling and visualizing test environments and its collision checkers and RRT~\citep{lavalle2000rapidly} implementation for motion planning. Since there has been very little research on the task and motion planning problem in stochastic settings, there are no standardized benchmarks.
We evaluated our algorithms by creating a hangar model in OpenRAVE for the aircraft inspection problem (Fig.\,\ref{fig:scenario}). UAV actions in this domain include actions for moving to various components of the aircraft, such as the left and right wings, nacelles, fuselage, etc. Each such action could result in the UAV reaching the specified component or a region around the component. The inspection action for a component had the stochastic effect of localizing a fault's location. The environment included docking stations that the UAV could reach and recharge on reserve battery power. Generators for concretizing all actions except the inspect action uniformly sampled  poses in the target regions. Some of these poses naturally lead to shorter trajectories and therefore lower battery usage, depending on the UAV's current pose. However, we used uniform-random samples to evaluate the performance of the algorithm while avoiding domain-specific enhancements. The generator for \emph{inspect($s$)} simulated an inspection pattern by randomly sampling five waypoint poses in an envelope around $s$ and ordering them along the medial axis of the component. We used a linear function of the trajectories to keep track of battery usage at the low level and to report insufficient battery as the \emph{failureReason} when infeasibility was detected. This function was used to provide failure reasons to the high-level when the battery level was found to be insufficient.

 Fig.\,\ref{fig:results_mdp} shows the performance of our
 approach for producing execution strategies with motion planning
 refinements as a function of the time for which the algorithm is
 allowed to run. The red lines show the number of nodes in the
 high-level policy that have been evaluated, refined, and potentially
 replaced with updated policies that permit low-level plans. The blue
 lines show the probability with which the policy available at any
 time during the algorithm's computation will be able to handle all
 possible execution-time outcomes. The different plots show how these relations change as we increase the level of uncertainty in the domain. The horizon is fixed at ten high-level decision epochs (each of which can involve arbitrarily long movements) and the number of parts with faults is fixed at two. The policy generated by LAO* is unrolled into a tree prior to the start of refinement. The reported times include the time taken for unrolling.

Our main result is that that our anytime algorithm balances complexity of computing task and motion policies with time very well and produces desirable concave anytime peformance profiles. Fig.\,\ref{fig:results_mdp} shows that when noise in the agent's actuators
and sensors is set at $5\%$, with $10\%$ of computation our algorithm computes an executable
policy that misses only the least likely $10\%$ of the possible execution outcomes. This policy is computed in less
than 10 seconds. In the worst case, with a 20\% error rate in actuators and sensors (sensors used in practice are much more reliable), we miss only about 20\% of the execution trajectories with 40\% of the computation.

%% file: related.tex
\section{Other Related Work}
\label{sec:related}
There has been a renewed interest in integrated task and motion planning algorithms. Most research in this direction has been focused on deterministic environments~\citep{cambon09_asymov,plaku10_sampling,dornhege12_semantic,kaelbling11_hierarchical,garrett15_ffrob,dantam16_incremental}. \cite{kaelbling13_hpnPOMDP} consider  a partially observable formulation of the problem. Their approach utilizes regression modules on belief fluents to develop a regression-based solution algorithm. \cite{sucan12_tmp_mdp} use an explicit multigraph to represent the plan or policy for which motion planning refinements are desired.  \cite{hadfield15_modular} address problems where the high-level formulation is deterministic and the low-level is determinized using most likely observations. In contrast, our approach employs abstraction to bridge MDP solvers and motion planners to solve problems where the high-level model is stochastic. In addition, the transitions in our MDP formulation depend on properties of the refined motion planning trajectories (e.g., battery usage). 

Principles of abstraction in MDPs have been well studied~\citep{hostetler14_state,bai16_markovian,li06_abstractMDP,singh95_abstractRL}. However, these directions of work assume that the full, unabstracted MDP can be efficiently expressed as a discrete MDP. \cite{marecki06_cmdp} consider continuous time MDPs with finite sets of states and actions. In contrast, our focus is on MDPs with high-dimensional, uncountable state and action spaces. Recent work on deep reinforcement learning  (e.g., \citep{hausknecht16_iclr,mnih15_drl}) presents  approaches for using deep neural networks in conjunction  with reinforcement learning to solve MDPs with continuous state spaces. We believe that these approaches can be used in a complementary fashion with our proposed approach. They could be used to learn maneuvers spanning shorter-time horizons, while our approach could be used to efficiently abstract their representations and to use them as actions or macros in longer-horizon tasks. 

Efforts towards improved representation languages are orthogonal to our contributions~\citep{fox02_pddl+}. The fundamental computational complexity results indicating growth in complexity with increasing sizes of state spaces, branching factors, and time horizons remain true regardless of the solution approach taken. It is unlikely that a uniformly precise model, a simulator at the level of precision of individual atoms, or even circuit diagrams of every component used by the agent will help it solve the kind of complex tasks on which humans would appreciate assistance. On the other hand, not using any model at all would result in dangerous agents that would not be able to safely evaluate the possible outcomes of their actions. Our results show that these divides can be bridged using hierarchical modeling and solution approaches that simplify the representational requirements and offer computational advantages that could make autonomous robots feasible in the real world.

%% file: conclusions.tex
\section{Conclusions}

Our experiments showed that
starting with an imprecise model, refining it based on the
information required to evaluate different courses of action is an
efficient approach for the synthesis of high-level policies that are consistent with constraints that may be imposed by aspects of the model that are more abstract or imprecise. While full models of realistic problems can overwhelm SDM solvers due to the uncountable branching factor and long time horizons, our hierarchical approach allows us to use SDM solvers while addressing more realistic problems involving physical agents.